\documentclass[runningheads]{llncs}
\usepackage[T1]{fontenc}
\usepackage{graphicx,verbatim}
\usepackage{caption}
\usepackage{subcaption}
\usepackage{booktabs}
\usepackage{multirow}
\usepackage{graphicx}
\usepackage{hyperref}
\usepackage{amssymb}
\usepackage{pifont}
\usepackage{breakcites}

\begin{document}
\title{Adaptable Segmentation Pipeline for Diverse Brain Tumors with Radiomic-Guided Subtyping and Lesion-Wise Model Ensemble}


%
\titlerunning{Adaptable MRI brain tumor segmentation pipeline}
\author{Daniel Capell\'{a}n-Mart\'{i}n\inst{1,2}*, Abhijeet Parida\inst{1,2}*, Zhifan Jiang\inst{1}*,\\ Nishad Kulkarni\inst{1}, Krithika Iyer\inst{1}, Austin Tapp\inst{1}, Syed Muhammad Anwar\inst{1,3},\\ Mar\'{i}a J. Ledesma-Carbayo\inst{2} and Marius George Linguraru\inst{1,3}\\\email{mlingura@childrensnational.org}}

\authorrunning{D. Capell\'{a}n-Mart\'{i}n, A. Parida, Z. Jiang et al.}
%
\institute{
Sheikh Zayed Institute for Pediatric Surgical Innovation, \\Children’s National Hospital, Washington, DC, USA 
\and 
Universidad Polit\'{e}cnica de Madrid and CIBER-BBN, ISCIII, Madrid, Spain
\and 
School of Medicine and Health Sciences, \\George Washington University, Washington, DC, USA
}

\maketitle              
\begin{abstract}
Robust and generalizable segmentation of brain tumors on multi‑parametric magnetic resonance imaging (MRI) remains difficult because tumor types differ widely. The BraTS 2025 Lighthouse Challenge benchmarks segmentation methods on diverse high-quality datasets of adult and pediatric tumors: multi-consortium international pediatric brain tumor segmentation (PED), preoperative meningioma tumor segmentation (MEN), meningioma radiotherapy segmentation (MEN-RT), and segmentation of pre- and post-treatment brain metastases (MET).
We present a flexible, modular, and adaptable pipeline that improves segmentation performance by selecting and combining state-of-the-art models and applying tumor‑ and lesion‑specific processing before and after training. Radiomic features extracted from MRI help detect tumor subtype, ensuring a more balanced training. Custom lesion‑level performance metrics determine the influence of each model in the ensemble and optimize post‑processing that further refines the predictions, enabling the workflow to tailor every step to each case.
On the BraTS testing sets, our pipeline achieved performance comparable to top-ranked algorithms across multiple challenges.
These findings confirm that custom lesion-aware processing and model selection yield robust segmentations yet without locking the method to a specific network architecture. Our method has the potential for quantitative tumor measurement in clinical practice, supporting diagnosis and prognosis.

* These authors contributed equally.
\keywords{Brain tumor segmentation \and MedNeXt \and Meningiomas \and Metastases \and MRI \and nnU-Net  \and Pediatric brain tumors \and Tumor subtyping}

\end{abstract}
%
%
%









\section{Introduction}

In the United States, cancer is the second leading cause of death overall and the primary cause among individuals younger than 85 years old. Brain and other central nervous system tumors are the top cause of cancer death in children and adolescents under 20, and brain tumors also lead cancer mortality in men aged 20–39. In 2025, about two million new cancer cases and 618,000 cancer-related deaths are projected in the U.S.~\cite{cancer-stats-2025}. Early, accurate diagnosis is essential to improve outcomes, but wide variation in tumor appearance across imaging devices and population makes consistent assessment difficult, highlighting the need for reliable quantitative diagnositc and prognostic tools.

In neuro-oncology, accurate segmentation of brain tumors in multi-parametric magnetic resonance imaging (mpMRI) is a fundamental step for diagnosis, treatment planning, and longitudinal monitoring. Yet manual contouring remains labor-intensive and prone to inter‑observer variability, motivating robust automated solutions in the clinical workflow. 

The international Brain Tumor Segmentation (BraTS) Challenge~\cite{brats2021,bakas1,bakas2,bakas3,medperf,brats2015}, organized in conjunction with the Medical Image Computing and Computer Assisted Intervention (MICCAI) conference, has driven and benchmarked segmentation algorithmic innovation since MICCAI 2012. In its 2025 edition, BraTS expanded into a suite of challegnes that cover a broader range of tumors and tasks, earning recognition as one of MICCAI’s three Lighthouse Challenges.

Recent deep learning models such as nnU‑Net~\cite{nnunet}, MedNeXt~\cite{mednext}, Swin UNETR~\cite{swinunetr2,swinunetr}, and emerging Mamba‑based hybrids~\cite{umamba,segmambav2} have defined the current state of the art (SOTA) in medical image segmentation. As the SOTA models have become more stable and robust across diverse tasks, their reported Dice scores 
have converged within narrow margins~\cite{nnunet_revisited}, signaling a performance plateau. While ever more complex architectures are possible, they would offer diminishing added values. 

BraTS‑winning solutions have demonstrated that model-independent and GPU-free heuristics, including ensemble voting~\cite{maani2023advanced}, size‑aware connected component filtering and ET‑to‑NCR relabelling~\cite{capellan2023model}, and adaptive refinement schemes~\cite{jiang2024magnetic,parida2024adult}, can substantially boost segmentation performance. Instead of pursuing ever‑\allowbreak deeper or more intricate networks, we believe that the next gains will arise from target-specific pipeline design. Robust pre‑processing such as radiomic‑guided fold splitting mitigates sampling bias; lesion‑aware ensemble weighting exploits complementary backbone strengths; and subtype‑specific post‑processing removes residual artifacts.

We therefore introduce and apply a flexible, modular, backbone‑independent pipeline to multiple BraTS 2025 Lighthouse Challenges to show its versatility. This end‑to‑end strategy consistently outperforms any single backbone, demonstrating that thoughtful pipeline engineering remains promising once model performance plateaus.

\section{Methods}
\subsection{Dataset}
Our segmentation pipeline was applied to four BraTS 2025 Lighthouse tasks: multi-consortium international pediatric brain tumor segmentation (PED; train = 261, val = 91)~\cite{PEDarxiv2024,PEDarxiv}, preoperative meningioma tumor segmentation (MEN; train = 1000, val = 141)~\cite{MENarxiv}, meningioma radiotherapy segmentation (MEN-RT; train = 500, val = 70)~\cite{MENarxiv2024}, and segmentation of pre- and post-treatment brain metastases (MET; train = 1296, val = 179)~\cite{METarxiv}. Each case contains co-registered isotropic pre-contrast T1-weighted (T1), contrast-enhanced T1-weighted (T1CE), T2-weighted (T2), and T2-weighted fluid-attenuated inversion recovery (T2-FLAIR) MRI sequences. In MEN-RT cohort, only T1CE is available and in its original image space. 

Reference standard annotations of the tumor sub-regions were created and approved by expert neuroradiologists. Concretely the following tumor annotations and subregions were defined for each of the Lighthouse challenges considered in this work. 

\begin{itemize}
    \item \textbf{PED:} enhancing tumor (ET), non-enhancing tumor (NET), cystic component (CC), peritumoral edema (ED), tumor core (TC=ET+NET+CC), and whole tumor (WT=TC+ED). 
    \item \textbf{MEN:} ET, non-enhancing tumor core (NETC), surrounding non-enhancing FLAIR hyperintensity (SNFH), TC (ET+NETC), and WT (TC+SNFH).
    \item \textbf{MEN-RT:} gross tumor volume (GTV).
    \item \textbf{MET:} ET, NETC, SNFH, resection cavity (RC; delineates the resection of region within the brain in post-treatment cases), TC (ET+NETC), and WT (TC+SNFH).
\end{itemize}
For more details, please refer to the challenge website: 
\url{https://www.synapse.org/Synapse:syn64153130/wiki/}.

\subsection{Segmentation Pipeline}
\begin{figure}
    \centering
    \includegraphics[width=\linewidth]{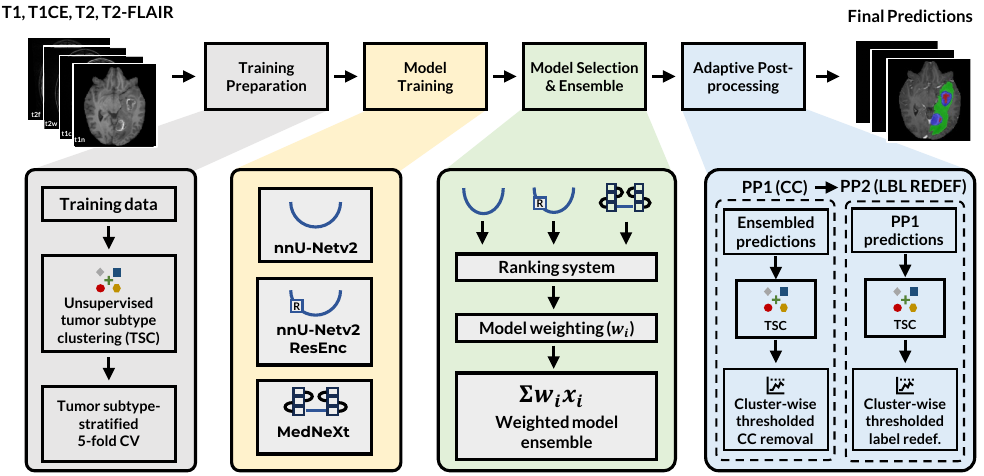}
    \caption{Overview of the segmentation pipeline. PP, CC, and LBL REDEF refer to post-processing, connected components, and label redefinition, respectively.}
    \label{fig:enter-label}
\end{figure}

Our pipeline uses MRI radiomic features and lesion‑wise metrics to adapt the workflow to each tumor subtype. As summarized in Fig~\ref{fig:enter-label}, it comprises four stages: (i) training data preparation: stratified fold splitting based on radiomic feature clustering; (ii) model training; (iii) model selection and ensembling; and (iv) adaptive post‑processing, including optimal removal of small connected components and label redefinition.

\subsection{Stratified Training Data Preparation}
In five-fold cross-validation (5-fold CV) training, stratified and balanced folds outperform purely random fold splits.  While stratification by tumor volume has been explored, we argue that radiomic features-capturing the full heterogeneity of tumor appearance-provide a more robust basis for fold assignment. For each case in the training set, we computed 14 shape- and 93 appearance-based radiomic features per MRI sequence based on WT mask provided by reference annotation, using PyRadiomics~\cite{van2017computational}, following the protocol of Jiang \textit{et~al.}~\cite{jiang2023automatic}.  

\begin{figure}[htbp]
    \centering
    \includegraphics[width=\linewidth]{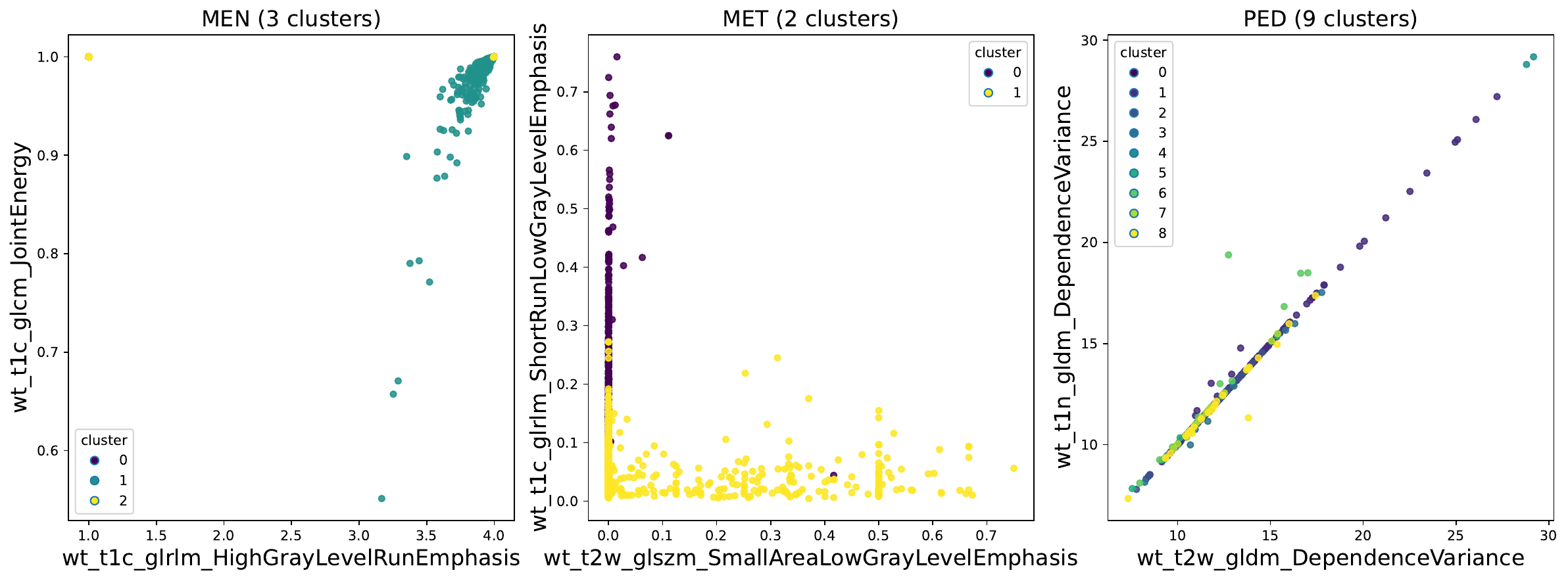}
    \caption{Clustering visualization across challenges using top two radiomic features.}
    \label{fig:clustering}
\end{figure}

To reduce the large number of radiomic features, we keep only the principal components that explain $90\%$ of the variance and then partition the cases with $k$-means clustering using reduced features~\cite{jiang2024enhancing,jiang2024magnetic}.  
The optimal number of clusters is determined by maximizing the silhouette coefficient in the training folds. Each cluster is randomly split into five folds, yielding the final CV sets for training.
Fig.~\ref{fig:clustering} illustrates the clustering based on the two radiomic features most relevant for cluster separation.

\subsection{Model Training}
Based on the reported performance in~\cite{nnunet_revisited} and our previous experience, we selected three state-of-the-art models: nnU-Net V2, nnU-Net with residual encoder (nnU-Net ResEnc M), and MedNeXt M (k=3, 17.6M parameters, 248 GFlops). Each model was trained using a 5-fold CV strategy with input image patches of 128 $\times$ 128 $\times$ 128 voxels. The nnU-Net variants and MedNeXt-M were trained using same settings, employing label-wise softmax activation, a class-weighted loss function combining Dice and cross-entropy losses, optimized by the stochastic gradient descent (SGD) with Nesterov momentum (initial learning rate=0.01, momentum=0.99, weight decay=3e-5), for 200 epochs on NVIDIA A100 (40 GB) GPUs.
The implementation is available through the official frameworks repositories: \href{https://github.com/MIC-DKFZ/nnUNet}{https://github.com/MIC-DKFZ/nnUNet} and 
\href{https://github.com/MIC-DKFZ/MedNeXt}{https://github.com/MIC-DKFZ/MedNeXt}.

\subsection{Label-wise Metrics as Objective Function F}
\label{subsec:F}
As model training relies on a loss function, each downstream stage (model selection, ensembling, and post‑processing) also needs a quantitative objective so the pipeline can be customized automatically to each tumor type and region. This objective function can simply be the Dice similarity coefficient (DSC). However, in reality or in the context of challenges, DSC alone is insufficient because segmentation performance is evaluated with several metrics, including DSC, normalized surface distance (NSD) and Hausdorff distance (HD), which sit on different scales and can not be added directly. Also, integrating scores across tumour sub‑regions (ET, TC, WT etc.) compounds another challenge.

We therefore adopt the ranking strategy proposed by BraTS~\cite{LaBella2025-oh}. For every candidate prediction obtained on the training set using 5-fold CV, we compute lesion‑wise metrics for all sub‑regions, rank the predictions case‑wise, and average these ranks across metrics and regions. The resulting internal CV score $\mathbf{F}$ is a single value (smaller is better) that combines all lesion-wise metrics, aligns with the official BraTS leaderboard and remains robust to outliers. Our implementation is open‑source and available at \href{https://github.com/Pediatric-Accelerated-Intelligence-Lab/BraTS-Unofficial-Ranker}{github.com/Pediatric-Accelerated-Intelligence-Lab/BraTS-Unofficial-Ranker}.

\subsection{Model Selection and Ensemble}
Each trained model $M_i$ is ranked by the approach in Section~\ref{subsec:F} with a score $F_i$. 
Then we used a weighted model ensemble strategy when adding the averaged 5-fold probability output from each model. The weight of each model is defined as $W_i = \sum_{j \neq i} F_j / \sum_{i} F_i$. The weights for each model are summarized in Table~\ref{tab:m_weights}.
\begin{table}
\caption{Weights for each model across tasks.}
    \centering
    \begin{tabular}{lccccc}
    \hline  
        && \textbf{PED}  & \textbf{MEN} & \textbf{MEN-RT} & \textbf{MET} \\
     \hline    
    $W_0$ (nnU-Net V2) & &0.332  & 0.323 & 0.336 & 0.295 \\
    $W_1$ (nnU-Net ResEnc M)  & &0.331  & 0.326  &  0.338 & 0.296 \\
    $W_2$ (MedNeXt M) & &0.337 & 0.351 & 0.326 & 0.409 \\
    \hline
    \end{tabular}
    
    \label{tab:m_weights}
\end{table}

We also evaluated the STAPLE~\cite{staple} ensemble strategy. It lagged behind the weighted ensemble when we used the three 5-fold-averaged predictions on the training data. However, STAPLE performed better on the validation set for MEN-RT task because treating each fold’s model as an independent candidate yielded 15 predictions, which markedly enhanced DSC, NSD, and rank for the MEN‑RT validation cohort.

\subsection{Adaptive Post-processing}
After ensembling, we recompute stratification clusters on the predicted masks, enabling cluster‑ and label‑specific thresholds for adaptive post‑processing.

\noindent\textbf{Post-processing for the removal of isolated connected components (PP1-CC):} To remove small isolated CCs that are likely false or noisy predictions, we perform a grid search across clusters and labels over volumes from 0 to 500 voxels in 25‑voxel increments. For each threshold, we filter the predicted mask and compute the internal ranking score, and finally select the threshold that achieves the best averaged rank. This cluster‑ and label-specific filter eliminates fragments that would otherwise be counted as false‑positive lesions.

\begin{figure}[htbp]
    \centering
    \includegraphics[width=\linewidth]{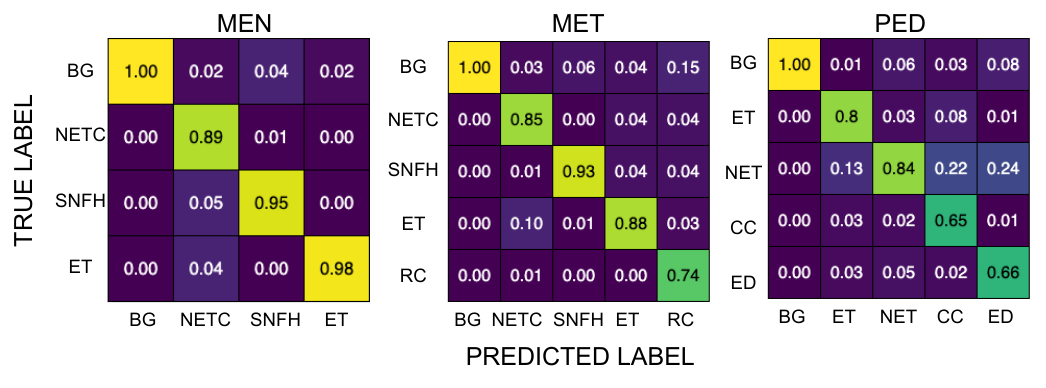}
    \caption{Post-PP1 confusion matrix indicating labels that need to be redefined in PP2. BG refers to background.}
    \label{fig:cm}
\end{figure}

\begin{figure}[htbp]
    \centering
    \includegraphics[width=\linewidth]{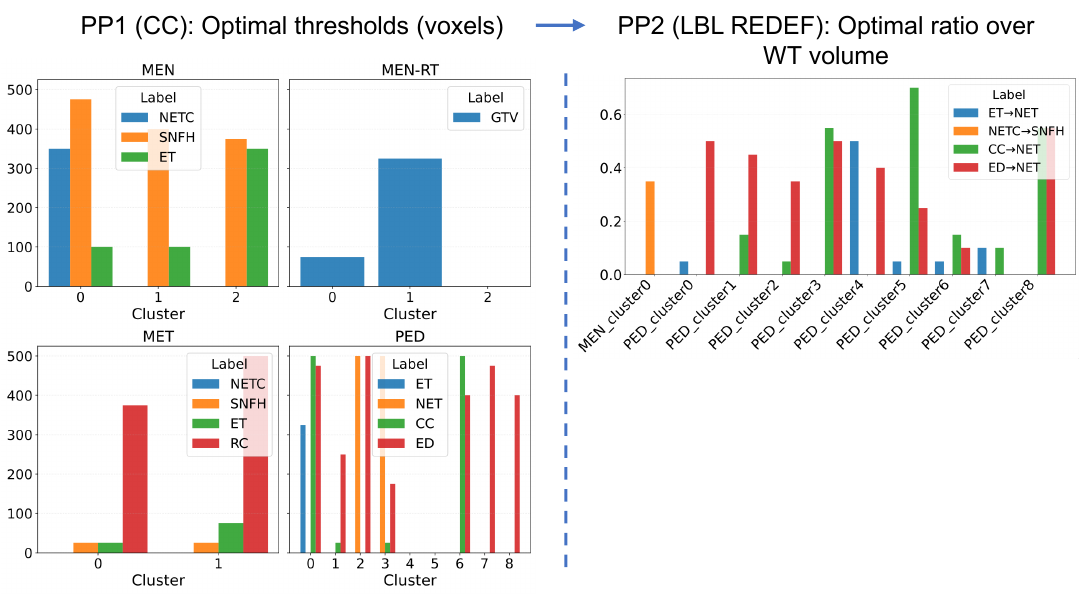}
    \caption{Adaptive Post-processing: optimal thresholds per cluster and label across challenges.}
    \label{fig:pp}
\end{figure}

\noindent\textbf{Post-processing for label redefinition (PP2-LBLREDEF):} A second adaptive PP fine‑tunes the consistency of tumor sub-regions (labels). We correct systematic label confusions in a data‑driven way. After PP1-CC, we build a confusion matrix (Fig.~\ref{fig:cm}) over all predicted masks again to identify pairs of frequently swapped labels. For every such pair (label$_x$,label$_y$), we search, within each cluster, for the threshold on the volume ratio label$_x$/WT that maximizes internal CV metrics (\textit{e.g.} rank). If a case with predicted mask falls below this threshold, all label$_x$ voxels are converted to label$_y$. This ratio‑based PP2-LBLREDEF enforces anatomically plausible label volumes and improves the performance of the BraTS metrics at the lesion level.  

At inference, a new test case is assigned to its nearest radiomic cluster, after model ensemble and the cluster‑specific post‑processing threshold values are then applied. Fig.~\ref{fig:pp} shows the optimal thresholds identified by CV-based grid search. 

    

\section{Results}
The evaluation of the model prediction on the validation set was performaed on the Synapse platform. The models were assessed for each of the tumor regions using the lesion-wise DSC and NSD with boundary threshold of 1.0 mm. 

\begin{figure}[htbp]
    \centering
    \includegraphics[width=\linewidth]{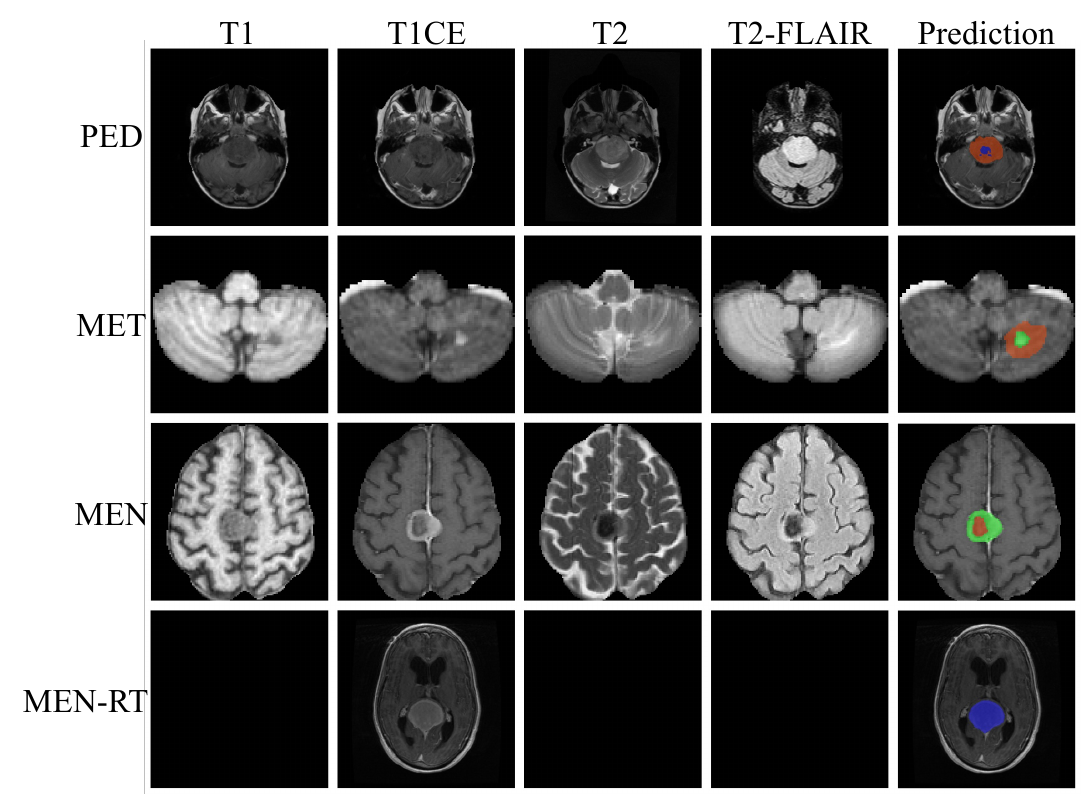}
    \caption{Qualitative results showing median lesion-wise Dice (L) or global Dice (G) of the whole tumor. PED: 0.979L, orange=NET, blue=ET; MET: 0.868G, orange=SNFH, green=ET; MEN: 0.961L, orange=SNFH, green=ET; MEN-RT: 0.887L, blue=GTV.}
    \label{fig:results}
\end{figure}

Quantitative results of our models across the validation and testing data for each challenge are shown in Tables ~\ref{tab:val-results-peds} and~\ref{tab:val-results-others}, for PED, MET, MEN, and MEN-RT, respectively. These evaluations were performed automatically by the challenge's digital platform, with no access to the reference standard annotations on the validation set and no access to any testing data including images and labels. 
Figure~\ref{fig:results} illustrated qualitative results on validation cases for PED, MET, MEN, and MEN-RT, respectively.

\begin{table}[h!]
\caption{\textbf{PED quantitative results} Lesion-wise (LW) Dice coefficients and Normalized Surface Distance (NSD) at threshold 1mm were computed for enhancing tumor (ET), tumor core (TC), whole tumor (WT), non-enhancing tumor (NET), cystic components(CC), and edema (ED), respectively.}
\centering
\resizebox{1.0\textwidth}{!}{%
\begin{tabular}{@{}clllccccccclcccccc@{}}
\toprule
\multirow{2}{*}{\textbf{Task}} &  & \multicolumn{1}{l}{\multirow{2}{*}{\textbf{Model}}} & \textbf{} & \multicolumn{6}{c}{\textbf{LW Dice}} & \textbf{\hspace{0.2cm}} & \multicolumn{6}{c}{\textbf{LW NSD thresh. 1mm}} \\ \cmidrule(l){4-17} 
 &  & \multicolumn{1}{c}{} & \textbf{} & \textbf{CC} & \textbf{ED} & \textbf{ET} & \textbf{NET}& \textbf{TC}& \textbf{WT}&\textbf{} & \textbf{CC} & \textbf{ED} & \textbf{ET} & \textbf{NET}& \textbf{TC}& \textbf{WT} \\ \midrule
&  & MedNeXt(M) & &0.735 & 0.857& 0.694 & 0.899& 0.926&0.927 & &0.749&0.857 &0.748 &0.878&0.903&0.905 \\
\textbf{PED} &  & nnU-Net & &0.713 & 0.813& 0.630 & 0.900& 0.928&0.928 & &0.762&0.813 &0.693 &0.885&0.914&0.915 \\
\textbf{Validation} &  & nnU-Net-ResEnc(M) & &0.719 & 0.857&0.692 & 0.902& 0.931&0.932 & &0.735&0.857 &0.755 &0.883&0.910&0.913 \\
\cmidrule(l){2-17} 
 \textbf{N=91} &  & Ensemble & &0.704 &0.868& 0.677 & 0.906& 0.933&0.933 & &0.736&0.868 &0.731 &0.891&0.916&0.917 \\
&  & Post-processing & &0.72 & 0.967& 0.65 & 0.907& 0.933&0.933 & &0.72&0.967 &0.696 &0.892&0.914&0.914 \\ 
\midrule
\textbf{PED} &  & Mean &  & 0.669 & 0.892 & 0.672 & 0.851 &0.908& 0.915 && 0.67 & 0.892 & 0.714 & 0.815 & 0.82 & 0.83 \\
\textbf{Testing} &  & Standard Deviation &  & 0.468 & 0.304 & 0.361 & 0.201 & 0.145 & 0.14 && 0.466 & 0.304 & 0.371 & 0.206 & 0.228 & 0.213 \\

\bottomrule
\end{tabular}%
}
\label{tab:val-results-peds}
\end{table}

\begin{table}[h!]
\caption{\textbf{MET, MEN, MEN-RT quantitative results}. Lesion-wise (LW) or global (G) Dice coefficients and Normalized Surface Distance (NSD) at 1mm-threshold were computed for enhancing tumor (ET), tumor core (TC), whole tumor (WT), resection cavity (RC), and gross tumor volume (GTV).}
\centering
\resizebox{1.0\textwidth}{!}{%
\begin{tabular}{@{}clllcccccclccccc@{}}
\toprule
\multirow{2}{*}{\textbf{Task}} &  & \multicolumn{1}{l}{\multirow{2}{*}{\textbf{Model}}} & \textbf{} & \multicolumn{5}{c}{\textbf{Dice}} & \textbf{\hspace{0.2cm}} & \multicolumn{5}{c}{\textbf{NSD thresh. 1mm}} \\ \cmidrule(l){4-15} 
 &  & \multicolumn{1}{c}{} & \textbf{} & \textbf{ET} & \textbf{TC} & \textbf{WT} & \textbf{RC}& \textbf{GTV}& \textbf{} & \textbf{ET} & \textbf{TC} & \textbf{WT} & \textbf{RC}& \textbf{GTV} \\ \midrule
&  & MedNeXt(M) & &0.747 & 0.768&0.776 & 0.628& & &0.588&0.595 &0.496 &0.591& \\
\textbf{MET} &  & nnU-Net & &0.731 & 0.752& 0.754 & 0.656& & &0.559 & 0.566 &0.468 &0.617 & \\
\textbf{Validation (G)} &  & nnU-Net-ResEnc(M) & &0.744 & 0.761& 0.766 & 0.693 & & & 0.570 & 0.577 &0.473 & 0.654 & \\
\cmidrule(l){2-15}
 \textbf{N=179} &  & Ensemble & &0.744 & 0.765& 0.769 & 0.705& & & 0.579 &0.587 & 0.489 & 0.668 & \\
&  & Post-processing & & 0.699 & 0.72 & 0.719 & 0.907 & & & 0.535 & 0.544 &0.442 &0.871 & \\
\midrule
\textbf{MET} &  & Mean &  & 0.544 & 0.555 & 0.561 & 0.86 & & & 0.606 & 0.607 & 0.589 & 0.858 & \\
\textbf{Testing (LW)} &  & Standard Deviation &  & 0.317 & 0.322 & 0.322 & 0.322 & & & 0.336 & 0.338 & 0.32 & 0.321 & \\
\midrule
&  & MedNeXt(M) & &0.841 & 0.863& 0.839 & & & &0.853&0.87 &0.842 && \\
\textbf{MEN} &  & nnU-Net & &0.811 & 0.820& 0.834 & & & &0.824&0.823 &0.832 && \\
\textbf{Validation (LW)} &  & nnU-Net-ResEnc(M) & &0.807 & 0.822& 0.825 & & & &0.818&0.825 &0.826 && \\
\cmidrule(l){2-15}
 \textbf{N=141} &  & Ensemble & &0.835 & 0.844& 0.836 & & & &0.847&0.850 &0.839 && \\
&  & Post-processing & & 0.856 & 0.855& 0.848 & & & & 0.873& 0.865 & 0.854 && \\ 
\midrule
\textbf{MEN} &  & Mean &  & 0.881 & 0.878 & 0.868 & & & & 0.885 & 0.878 & 0.86 & & \\
\textbf{Testing (LW)} &  & Standard Deviation &  & 0.224 & 0.226 & 0.224 & & & & 0.218 & 0.221 & 0.216 & & \\
\midrule
&  & MedNeXt(M)  & & & &  & & 0.804& && & &&0.671 \\
\textbf{MEN-RT} &  & nnU-Net & & & &  & & 0.764& && & &&0.618 \\
\textbf{Validation (LW)} &  & nnU-Net-ResEnc(M)  & & & &  & & 0.795& && & &&0.651 \\
\cmidrule(l){2-15} 
 \textbf{N=70} &  & Ensemble  & & & &  & & 0.796& && & &&0.654 \\
&  &Post-processing  & & & &  & & 0.796 & && & &&0.654 \\ 
\midrule
\textbf{MEN-RT} &  & Mean & & & & & & 0.807 & & & & & & 0.683 \\
\textbf{Testing (LW)} &  & Standard Deviation & & & & & & 0.201 & & & & & & 0.236 \\
\bottomrule
\end{tabular}%
}
\label{tab:val-results-others}
\end{table}

\section{Discussion}



Our experiments confirm that carefully engineered pre‑ and post‑processing can still deliver tangible gains even when base architectures have plateaued. Notably, some gains in Dice or NSD were limited on validation set but became evident under 5-fold CV, underscoring the danger of optimizing solely on a limited and sometimes skewed validation set.

Because the training set is much larger than the validation set, the latter can present a skewed tumor‑type distribution and thus provide an unreliable signal for hyper‑parameter and threshold tuning. We therefore recommend selecting models and thresholds by CV on the training set, using validation scores only as a sanity check. In the same spirit, the internal rank metric—aggregating Dice and NSD across tumor regions, proved more robust than any single metric, especially when candidate models were numerous. However, its sensitivity to the number of candidates (rank scores can be very close if only three candidates) suggests that alternative weighting schemes merit exploration.

We previously tested that stratified fold splitting based on radiomic clusters yielded higher CV scores than random splits, indicating that heterogeneity‑aware sampling reduces overfitting. Additional ablations on multiple BraTS tasks are planned to quantify this effect. We also found that STAPLE fusion benefited the MEN‑RT task when each fold was treated as a separate candidate, indicating that larger model pools can regularize ensembles.

Limitations include our reliance on simple volume thresholds for PP1 and PP2. Future work will replace these heuristics with radiomic‑driven criteria analogous to those used for stratification, further tailoring post‑processing to both tumor morphology and appearance. 

Open‑sourcing our ranker and keeping all components model-independent, we aim to make the pipeline easy to deploy in clinical workflows. When its utility is evaluated in more clinical studies, the pipeline would have the potential for larger clinical impact beyond BraTS benchmarks. 

Finally, to facilitate reproducibility and extend the utility of our approaches, we have made the complete pipeline publicly available as easy-to-use Docker containers and a webapp for all the tasks. 
The Docker images are hosted at: \url{https://hub.docker.com/r/aparida12/brats2025} and the webapp is accessible at:  \url{https://segmenter.hope4kids.io/}.

\section{Conclusion}
This work introduces a tumor- and lesion‑aware ensemble pipeline that consistently segments four distinct brain tumour cohorts in the BraTS 2025 Lighthouse Challenge. By selecting and weighting models according to lesion‑level accuracy and adding subtype‑specific processing, the framework achieves strong segmentation performance relying on state-of-the-art network backbones. Its modular design makes it easy to adopt and extend, while the resulting quantitative tumor measurements could support clinical diagnosis, treatment planning, and longitudinal monitoring.

\begin{credits}
\subsubsection{\ackname} This work was supported by the National Cancer Institute (UG3 CA236536), the Spanish  Ministerio de Ciencia e Innovación, the Agencia Estatal de Investigación, NextGenerationEU grants PDC2022-133865-I00 and PID2022-141493OB-I00, and the EUCAIM project co-funded by the European Union (Grant Agreement \#101100633). The authors acknowledge the Universidad Politécnica de Madrid for providing computing resources on the Magerit Supercomputer. This work was also supported by the Comunidad de Madrid, Spain through the MAGERIT-CM project (TEC-2024/COM-44).

\end{credits}
%
%
%
\bibliographystyle{splncs04}
\bibliography{references}
\end{document}